\documentclass[10pt,journal,compsoc]{IEEEtran}
\ifCLASSOPTIONcompsoc
  \usepackage[nocompress]{cite}
\else
  \usepackage{cite}
\fi
\usepackage{hyperref} 
\hypersetup{
    colorlinks=true,
    linkcolor=red,
}
\usepackage{amssymb}
\usepackage{bbding}

\usepackage{booktabs}
\usepackage{tabularx}
\usepackage{array}

\usepackage{forest}
\usepackage{xcolor}
\usepackage{hyperref}

\usepackage{stfloats}
\usepackage{color}
\usepackage[dvipsnames]{xcolor}

\usepackage{multirow}

\usepackage{caption}
\usepackage{subcaption}
\usepackage{amsmath}

\usepackage{amssymb}
\usepackage{pifont}

\begin{document}
%
\title{
A Mechanistic View on Video Generation as World Models: State and Dynamics
}
\author{Luozhou~Wang$^{*}$,
        Zhifei~Chen$^{*}$,
        Yihua~Du,
        Dongyu~Yan,
        Wenhang~Ge,
        Guibao~Shen,
        Xinli~Xu,
        Leyi~Wu,
        Man~Chen,
        Tianshuo~Xu,
        Peiran~Ren,
        Xin~Tao,
        Pengfei~Wan,
        and~Ying-Cong~Chen$^{\dagger}$%
\IEEEcompsocitemizethanks{
    \IEEEcompsocthanksitem $^{*}$ Luozhou Wang and Zhifei Chen contributed equally to this work.
    \IEEEcompsocthanksitem L. Wang, Z. Chen, Y. Du, D. Yan, W. Ge, G. Shen, X. Xu, L. Wu, M. Chen, T. Xu, and Y.-C. Chen are with the Hong Kong University of Science and Technology (Guangzhou), Guangzhou, China.
    \IEEEcompsocthanksitem P. Ren is with Tongji University, Shanghai, China.
    \IEEEcompsocthanksitem X. Tao and P. Wan are with Kuaishou Technology, Beijing, China.
    \IEEEcompsocthanksitem Corresponding author: Ying-Cong Chen (yingcongchen@hkust-gz.edu.cn).
}
}

\IEEEtitleabstractindextext{%
\begin{abstract} Large-scale video generation models have demonstrated emergent physical coherence, positioning them as potential \textit{world models}. However, a gap remains between contemporary ``stateless'' video architectures and classic state-centric world model theories. This work bridges this gap by proposing a novel taxonomy centered on two pillars: \textbf{State Construction} and \textbf{Dynamics Modeling}. We categorize state construction into \textit{implicit} paradigms (context management) and \textit{explicit} paradigms (latent compression), while dynamics modeling is analyzed through \textit{knowledge integration} and \textit{architectural reformulation}. Furthermore, we advocate for a transition in evaluation from visual fidelity to functional benchmarks, testing physical persistence and causal reasoning. We conclude by identifying two critical frontiers: enhancing persistence via data-driven memory and compressed fidelity, and advancing causality through latent factor decoupling and reasoning-prior integration. By addressing these challenges, the field can evolve from generating visually plausible videos to building robust, general-purpose world simulators.
\end{abstract}

}

\maketitle

\IEEEdisplaynontitleabstractindextext

%
\IEEEpeerreviewmaketitle

\section{Introduction}
\label{sec:introduction}

\IEEEPARstart{T}{he} field of video generation has witnessed a paradigm shift in recent years. 
Driven by the development of large-scale video diffusion transformer models \cite{gupta2024photorealistic,lu2023vdt,ma2024latte,fan2025vchitect,liu2025lumina,zhang2025waver,wan2025wan,chen2025skyreels,ma2025step,agarwal2025cosmos,hacohen2024ltx,kong2024hunyuanvideo,genmo2025blog,openai2024sora,lin2024open,zheng2024open,yang2024cogvideox,BarTal2024LumiereAS,Chen2023SEINESV,veo2025,kuaishou2024klingai,gao2025seedance,hailuoai2025,Bruce2024GenieGI,pikalabs2024pika}, the capability of generative models has evolved from producing short, low-resolution clips to rendering cinematic-quality sequences with unprecedented temporal consistency. 
Leading models such as Sora~\cite{openai2024sora}, Veo~\cite{veo2025}, Kling~\cite{kuaishou2024klingai}, Wan~\cite{wan2025wan}, and Gen-3~\cite{runway2024gen3} have demonstrated implications that extend far beyond mere visual fidelity. 
These models exhibit emergent physical coherence. 
For example, they learned to respect gravity, understand collision dynamics, and recognize object permanence without explicit physical instruction. 
As a result, video generation models are increasingly discussed not merely as content creation tools, but as potential \textbf{world models}~\cite{openai2024sora,hu2023gaia,veo2025,du2023learning, huang2025towards,li2025worldmodelbench,kang2024far,ren2025videoworld} that appear to simulate the physical evolution of the environment.

To critically assess this potential, it is necessary to revisit the theoretical origins of the concept of the ``world model.'' 
The evolution of world models has spanned multiple epochs—moving from cognitive science and control theory to the modern era of deep learning. 
The term has its roots in cognitive science, specifically in the theory of mental models \cite{craik1943nature, gentner2014mental, johnson1983mental}. 
These foundational works posited that intelligent organisms maintain a ``small-scale model'' of external reality to anticipate events, reason about counterfactuals, and simulate consequences before taking action.

Subsequently, in the field of control theory~\cite{kalman1960new, aastrom2021feedback}, this internal simulation process was explicitly formalized through mathematical rules. 
This transition defined the concept of an internal state by manually selecting key variables to represent the environment. 
In this framework, a transition equation—derived from physical laws or established rules—is constructed to evolve the state during inference.
In contrast an observation equation is used to map these states back to sensory inputs.

The concept has since evolved into the modern era of deep learning, with Model-Based Reinforcement Learning (MBRL) \cite{forrester1971counterintuitive, ha2018world, ha2018recurrent, hafner2019learning, hafner2020mastering, hafner2023mastering, hafner2025training, mnih2015human, mnih2016asynchronous, barth2018distributed, schrittwieser2020mastering, oh2017value, silver2017predictron, hansen2022temporal, hansen2023td, lee2023dreamsmooth, mattes2023hieros, liu2025continual, prasanna2024dreaming, hao2025neural} serving as one of the most representative methodologies. 
In these settings, the world model is a parameterized, data-driven system inextricably coupled with a policy agent in a closed-loop interaction. By learning from data, the model captures environmental laws, allowing the agent to "dream" or practice in a latent imagination space before execution in the real world.

All these discussions regarding world models can be synthesized into three core elements: observation, state, and dynamics. 
Observation refers to the perceptible modalities or raw data from the environment. 
The state, while lacking a singular universal definition, generally refers to a set of variables that provides a sufficient representation of the world by removing redundancies irrelevant to the current task. 
Dynamics is responsible for state transitions governed by underlying causal relationships.

Taking MBRL as a primary example, these approaches typically rely on an explicit State-Space Model (SSM). By constructing such explicit state representations, these models can achieve long-range reasoning with constant computational overhead. Furthermore, this structure facilitates explicit causal decoupling, which effectively isolates internal state transitions from the influence of external control inputs.

However, video models operate very differently from the world models mentioned above.
First, the training and inference of video models do not involve reinforcement learning or policy models. 
Instead, they work in an open-loop manner, learning by passively observing large amounts of data~\cite{ha2018world, huang2025towards,li2025worldmodelbench,kang2024far,ren2025videoworld}. 
Most video models~\cite{gupta2024photorealistic,lu2023vdt,ma2024latte,fan2025vchitect,liu2025lumina,zhang2025waver,wan2025wan,chen2025skyreels,ma2025step,agarwal2025cosmos,hacohen2024ltx,kong2024hunyuanvideo,genmo2025blog,openai2024sora,lin2024open,zheng2024open,yang2024cogvideox,BarTal2024LumiereAS,Chen2023SEINESV,veo2025,kuaishou2024klingai,gao2025seedance,hailuoai2025,Bruce2024GenieGI,pikalabs2024pika} use the transformer architecture, which models observation sequences directly.
Because of its high parallelism and efficiency, this architecture scales easily with more data and computing power, leading to very strong emergent capabilities.
However, these transformer-based models lack explicit state modeling.
Because they do not construct a hidden state to represent the world, they must keep a very large context window during long-term reasoning. 
This creates a significant memory and computational burden as the video gets longer.
Finally, the lack of an explicit state means that causal decoupling is not clearly achieved. 
Some video models~\cite{yang2024cogvideox,wan2025wan,ren2025cosmos,kong2024hunyuanvideo,lin2024open,zheng2024open} use bidirectional attention, which causes them to lose the ability to perform causal inference along a sequence. 
In addition, external inputs, such as text prompts, can interfere with the model’s understanding of causality. 
It becomes unclear whether the model makes an apple fall because it understands the scene's physical dynamics, or simply because the text prompt told it to.

Regarding the two aspects mentioned above—state and dynamics—this survey provides a detailed discussion on how current technical approaches address the challenges of causal reasoning and long-range inference in video models. Our core contributions are threefold:

\begin{itemize}
    \item \textbf{From Video Model to World Model:} We bridge the gap between high-level theory and practical implementation. Starting from the definition of a complete world model, we narrow our focus to the specific context of video generation, defining the position of video models within the broader framework of world models.
    
    \item \textbf{Novel Taxonomy:} We propose a novel taxonomy based on the key constituents of a world model: \textbf{State} and \textbf{Dynamics}.
    For state, we categorize existing literature into two paradigms: \textit{Implicit State}, where history is managed via context windows (e.g., KV Cache), and \textit{Explicit State}, where history is compressed into compact variables akin to traditional state-space models.
    For dynamics, we categorize research into \textit{Causal Knowledge Integration}, which combines video models with external, reasoning-intensive models to improve inference, and \textit{Causal Architecture Reformulation}, which redesigns the video model architecture to support causal reasoning inherently.
    
    \item \textbf{Evaluation and Future Work:} We establish a three-level classification for evaluating video models as world models. We summarize the evaluation of generated content across three stages: basic quality, physical persistence, and causality. Finally, we highlight future directions, focusing on how video models can better solve the problems of persistence and causal consistency.

\end{itemize}

\begingroup

\definecolor{paired-light-blue}{RGB}{198, 219, 239}
\definecolor{paired-dark-blue}{RGB}{49, 130, 188}
\definecolor{paired-light-orange}{RGB}{251, 208, 162}
\definecolor{paired-dark-orange}{RGB}{230, 85, 12}
\definecolor{paired-light-green}{RGB}{199, 233, 193}
\definecolor{paired-dark-green}{RGB}{49, 163, 83}
\definecolor{paired-light-purple}{RGB}{218, 218, 235}
\definecolor{paired-dark-purple}{RGB}{117, 107, 176}
\definecolor{paired-light-gray}{RGB}{217, 217, 217}
\definecolor{paired-dark-gray}{RGB}{99, 99, 99}
\definecolor{paired-light-yellow}{RGB}{231, 204, 100}

\forestset{
  outline scheme/.style={
    for tree={
      grow'=0,
      anchor=west,
      child anchor=west,
      parent anchor=east,
      align=left,
      rounded corners=2pt,
      draw,
      edge={-},
      l sep=10pt,
      s sep=6pt,
      inner sep=2pt,
      text width=3.6cm,
      font=\small,
    },
    before typesetting nodes={
      for tree={
        edge path'={(!u.parent anchor) -- ++(8pt,0) |- (.child anchor)}
      }
    },
    where level=1{
      text width=4.2cm,
    }{},
    where level=2{
      text width=4.8cm,
    }{},
    where level=3{
      text width=4.0cm,
    }{},
    where level=4{
      text width=5.1cm,
    }{},
  }
}
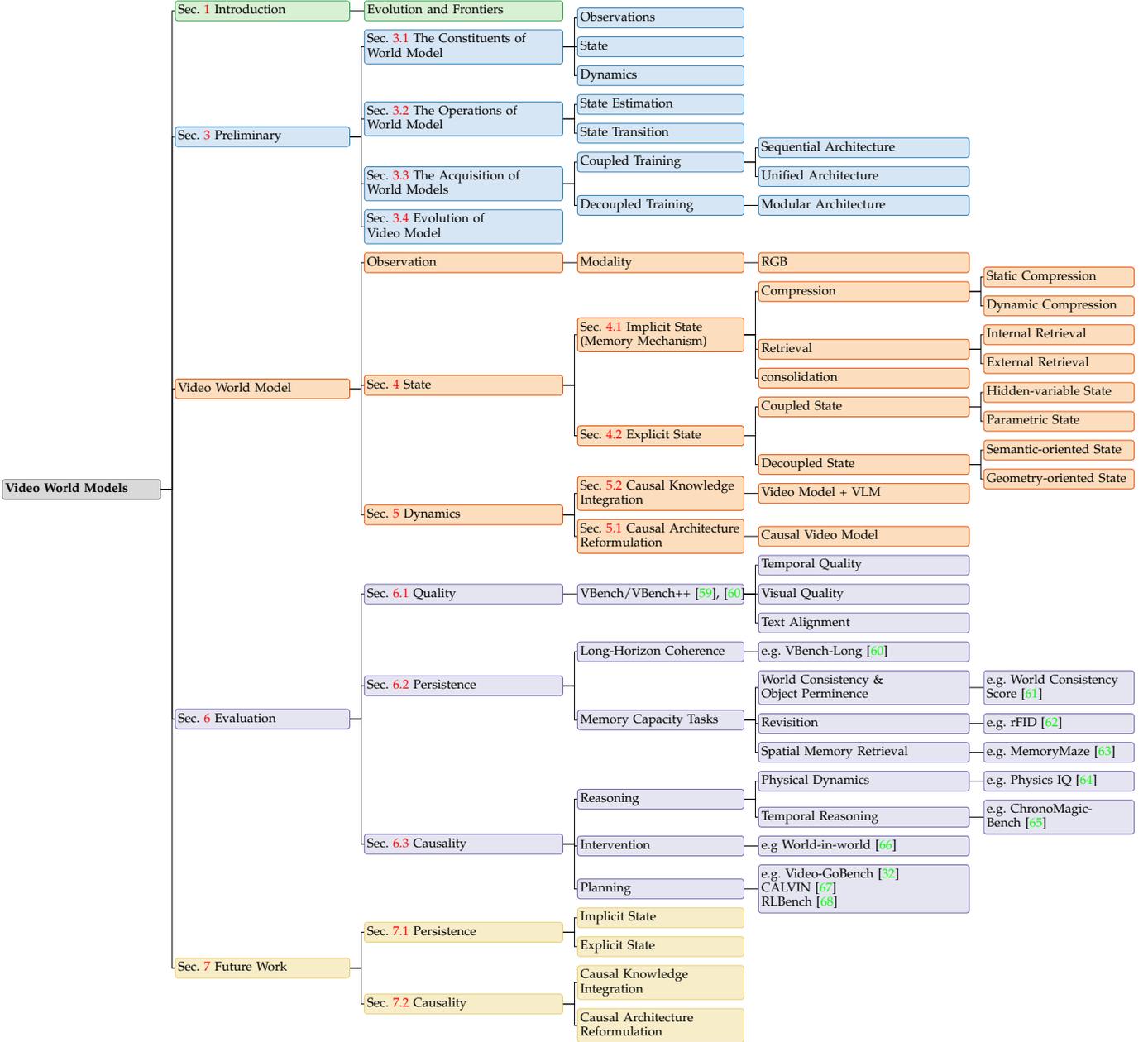
\begin{figure*}[t]
\centering
\resizebox{\textwidth}{!}{%
\begin{forest}
outline scheme
[{\bfseries Video World Models},
  fill=paired-light-gray, draw=paired-dark-gray, text width=3.8cm
  [Sec.~\ref{sec:introduction} Introduction,
    fill=paired-light-green!70, draw=paired-dark-green
    [Evolution and Frontiers,
      fill=paired-light-green!70, draw=paired-dark-green
    ]
  ]
  [Sec.~\ref{sec:preliminary} Preliminary,
    fill=paired-light-blue!70, draw=paired-dark-blue
    [Sec.~\ref{sec:pre_constituents} The Constituents of \\ World Model,
      fill=paired-light-blue!70, draw=paired-dark-blue
      [Observations, fill=paired-light-blue!70, draw=paired-dark-blue]
      [State,        fill=paired-light-blue!70, draw=paired-dark-blue]
      [Dynamics,     fill=paired-light-blue!70, draw=paired-dark-blue]
    ]
    [Sec.~\ref{sec:pre_operations} The Operations of \\ World Model,
      fill=paired-light-blue!70, draw=paired-dark-blue
      [State Estimation,
        fill=paired-light-blue!70, draw=paired-dark-blue
      ]
      [State Transition,
        fill=paired-light-blue!70, draw=paired-dark-blue
      ]
    ]
    [Sec.~\ref{sec:pre_acquisition} The Acquisition of \\ World Models,
      fill=paired-light-blue!70, draw=paired-dark-blue
      [Coupled Training,
        fill=paired-light-blue!70, draw=paired-dark-blue
        [Sequential Architecture, fill=paired-light-blue!70, draw=paired-dark-blue]
        [Unified Architecture,    fill=paired-light-blue!70, draw=paired-dark-blue]
      ]
      [Decoupled Training,
        fill=paired-light-blue!70, draw=paired-dark-blue
        [Modular Architecture, fill=paired-light-blue!70, draw=paired-dark-blue]
      ]
    ]
    [Sec.~\ref{sec:pre_evolution} Evolution of \\ Video Model,
      fill=paired-light-blue!70, draw=paired-dark-blue
    ]
  ]
  [Video World Model,
    fill=paired-light-orange!70, draw=paired-dark-orange
    [Observation,
      fill=paired-light-orange!70, draw=paired-dark-orange
      [Modality,
        fill=paired-light-orange!70, draw=paired-dark-orange
        [RGB,   fill=paired-light-orange!70, draw=paired-dark-orange]
      ]
    ]
    [Sec.~\ref{sec:videowm_state} State,
      fill=paired-light-orange!70, draw=paired-dark-orange
      [Sec.~\ref{sec:videowm_state_implicit} Implicit State \\ (Memory Mechanism),
        fill=paired-light-orange!70, draw=paired-dark-orange
        [Compression,
          fill=paired-light-orange!70, draw=paired-dark-orange
          [Static Compression,   fill=paired-light-orange!70, draw=paired-dark-orange]
          [Dynamic Compression, fill=paired-light-orange!70, draw=paired-dark-orange]
        ]
        [Retrieval,
          fill=paired-light-orange!70, draw=paired-dark-orange
          [Internal Retrieval, fill=paired-light-orange!70, draw=paired-dark-orange]
          [External Retrieval, fill=paired-light-orange!70, draw=paired-dark-orange]
        ]
        [consolidation, fill=paired-light-orange!70, draw=paired-dark-orange]
      ]
      [Sec.~\ref{sec:videowm_state_explicit} Explicit State,
        fill=paired-light-orange!70, draw=paired-dark-orange
        [Coupled State,
          fill=paired-light-orange!70, draw=paired-dark-orange
          [Hidden-variable State, fill=paired-light-orange!70, draw=paired-dark-orange]
          [Parametric State,      fill=paired-light-orange!70, draw=paired-dark-orange]
        ]
        [Decoupled State,
          fill=paired-light-orange!70, draw=paired-dark-orange
          [Semantic-oriented State, fill=paired-light-orange!70, draw=paired-dark-orange]
          [Geometry-oriented State, fill=paired-light-orange!70, draw=paired-dark-orange]
        ]
      ]
    ]
    [Sec.~\ref{sec:videowm_dynamics} Dynamics,
      fill=paired-light-orange!70, draw=paired-dark-orange
      [Sec.~\ref{sec:videowm_dynamics_integration} Causal Knowledge\\Integration,
        fill=paired-light-orange!70, draw=paired-dark-orange
        [Video Model + VLM, fill=paired-light-orange!70, draw=paired-dark-orange]
      ]
      [Sec.~\ref{sec:videowm_dynamics_reformulation} Causal Architecture\\Reformulation,
        fill=paired-light-orange!70, draw=paired-dark-orange
        [Causal Video Model, fill=paired-light-orange!70, draw=paired-dark-orange]
      ]
    ]
  ]
  [Sec.~\ref{sec:eval} Evaluation,
    fill=paired-light-purple!70, draw=paired-dark-purple
    [Sec.~\ref{sec:eval_quality} Quality,
      fill=paired-light-purple!70, draw=paired-dark-purple
      [VBench/VBench++~\cite{huang2023vbench,huang2024vbench++},
        fill=paired-light-purple!70, draw=paired-dark-purple
        [Temporal Quality, fill=paired-light-purple!70, draw=paired-dark-purple]
        [Visual Quality,   fill=paired-light-purple!70, draw=paired-dark-purple]
        [Text Alignment,   fill=paired-light-purple!70, draw=paired-dark-purple]
      ]
    ]
    [Sec.~\ref{sec:eval_persistence} Persistence,
      fill=paired-light-purple!70, draw=paired-dark-purple
      [Long-Horizon Coherence, fill=paired-light-purple!70, draw=paired-dark-purple
        [e.g. VBench-Long~\cite{huang2024vbench++}, fill=paired-light-purple!70, draw=paired-dark-purple]
      ]
      [Memory Capacity Tasks, fill=paired-light-purple!70, draw=paired-dark-purple
        [World Consistency \& \\Object Perminence, fill=paired-light-purple!70, draw=paired-dark-purple
            [e.g. World Consistency \\Score~\cite{rakheja2025worldconsistencyscore}, fill=paired-light-purple!70, draw=paired-dark-purple]
        ]
        [Revisition, fill=paired-light-purple!70, draw=paired-dark-purple
            [e.g. rFID~\cite{heusel2018rFID}, fill=paired-light-purple!70, draw=paired-dark-purple]
        ]
        [Spatial Memory Retrieval, fill=paired-light-purple!70, draw=paired-dark-purple
            [e.g. MemoryMaze~\cite{pasukonis2022maze}, fill=paired-light-purple!70, draw=paired-dark-purple]
        ]
      ]
    ]
    [Sec.~\ref{sec:eval_causality} Causality,
      fill=paired-light-purple!70, draw=paired-dark-purple
      [Reasoning,          fill=paired-light-purple!70, draw=paired-dark-purple
        [Physical Dynamics, fill=paired-light-purple!70, draw=paired-dark-purple
            [e.g. Physics IQ~\cite{motamed2025physiq},       fill=paired-light-purple!70, draw=paired-dark-purple]
        ]
        [Temporal Reasoning, fill=paired-light-purple!70, draw=paired-dark-purple
            [e.g. ChronoMagic-\\Bench~\cite{yuan2024chronomagic}, fill=paired-light-purple!70, draw=paired-dark-purple]
        ]
      ]
      [Intervention,       fill=paired-light-purple!70, draw=paired-dark-purple
        [e.g World-in-world~\cite{zhang2025world},   fill=paired-light-purple!70, draw=paired-dark-purple]
      ]
      [Planning,           fill=paired-light-purple!70, draw=paired-dark-purple
        [e.g. Video-GoBench~\cite{ren2025videoworld} \\ CALVIN~\cite{mees2022calvin} \\RLBench~\cite{james2019rlbench},   fill=paired-light-purple!70, draw=paired-dark-purple]
      ]
    ]
  ]
  [Sec.~\ref{sec:future} Future Work,
    fill=paired-light-yellow!35, draw=paired-light-yellow!90
    [Sec.~\ref{sec:future_persistence} Persistence,
      fill=paired-light-yellow!35, draw=paired-light-yellow!90
      [Implicit State, fill=paired-light-yellow!35, draw=paired-light-yellow!90]
      [Explicit State, fill=paired-light-yellow!35, draw=paired-light-yellow!90]
    ]
    [Sec.~\ref{sec:future_causality} Causality,
      fill=paired-light-yellow!35, draw=paired-light-yellow!90
      [Causal Knowledge \\Integration, fill=paired-light-yellow!35, draw=paired-light-yellow!90]
      [Causal Architecture \\Reformulation, fill=paired-light-yellow!35, draw=paired-light-yellow!90]
    ]
  ]
]
\end{forest}%
} 
\vspace{-4pt}
\caption{Overview of the paper structure following the mindmap ``From Video Generation Model to World Model''.}
\vspace{-12pt}
\label{fig:paper_structure_world_model}
\end{figure*}

\endgroup

\section{Related Survey}\label{sec:related_survey}

Research on memory mechanisms for Large Language Models (LLMs) has transitioned from early explorations of in-context learning and knowledge editing~\cite{zhao2023survey, wang2024knowledge} to highly systematized, cognitive-inspired architectural theories. A seminal milestone in this evolution was the survey by Zhang et al.~\cite{zhang2025survey} on agentic memory, which established a taxonomy of memory sources, forms, and operations to enable self-evolving capabilities in agents. In the 2024- 2025 period, the field saw an explosion of systematic frameworks. For instance, the survey by Wu et al.~\cite{wu2025human} introduced the 3D-8Q taxonomy mapping human cognitive stages to AI memory quadrants across personal, system, and temporal dimensions. Similarly, the study by Jia et al.~\cite{jiaai} delineated implicit, explicit, and agentic memory paradigms. In contrast, the work of Du et al.~\cite{du2025rethinking} defined the atomic dynamics of memory, such as consolidation and forgetting. Furthermore, recent surveys have addressed critical engineering and governance issues, including KV-cache optimization for long-context efficiency~\cite{li2024kvcache}, memory governance for "verifiable forgetting"~\cite{zhang2025memory}, personalized preference learning~\cite{liu2025personalized}, and multimodal retrieval-augmented generation~\cite{askany2025multimodal}.

Regarding the distinction between LLM (Text) and Video memory mechanisms: While text-based LLM memory primarily focuses on maintaining semantic consistency and factual accuracy across 1D sequences using RAG or KV-cache management~\cite{li2024kvcache, zhao2023survey}, video memory must model high-dimensional spatial-temporal dynamics and manage extreme signal redundancy~\cite{he2024ma, tang2025video}. Video LLMs typically employ streaming encoding or hierarchical memory banks to iteratively compress massive frames into compact representations~\cite{he2024ma}. This necessitates adaptive memory selection to pinpoint specific temporal "moments" in sequences spanning hours, rather than relying solely on the semantic vector matching common in text retrieval~\cite{tang2025video}.
\section{Preliminaries}\label{sec:preliminary}
This section transitions from the motivational overview in the Introduction to a more rigorous theoretical framework. We first establish a formal definition of a world model and subsequently detail its fundamental components, core operational functions, and the primary paradigms through which such models are acquired.

\subsection{The Constituents of World Models: Observation, State, and Dynamics} \label{sec:pre_constituents}
The conceptual foundation of the World Model is deeply rooted in cognitive science~\cite{johnson1983mental}, where it is characterized as a simplified internal mental representation of reality constructed within the human psyche.

By perceiving external sensory manifestations (\textbf{observations}) and distilling them into abstract internal concepts (\textbf{state}), humans utilize these internal representations to simulate causal outcomes (\textbf{dynamics}) within a mental ``sandbox.'' 
This internal simulation allows individuals to anticipate the potential consequences of various behaviors before executing physical actions, thereby serving as the fundamental cognitive substrate for proactive decision-making and strategic planning.

Beyond cognitive science, a closely related formalization of internal mental simulation emerged in control theory, where the behavior of the system being studied is modeled through a state-space representation~\cite{kalman1960new, aastrom2021feedback}. 
In this paradigm, the core variables summarizing the system's status are selected and maintained as the internal \textbf{state} $x$, which can fully describe the system without redundancy.
Its \textbf{dynamics} $\dot{x}$ (the derivative of $x$) and \textbf{observation}  $y$ can be derived using differential equations, which are artificially predefined using physical laws:
\begin{equation}
    \dot{x} = f(x, u), \quad y = g(x),
\end{equation}
where $u$ denotes the control input.
This state-space formulation provides a mathematically rigorous definition of the mappings from the \textbf{state} to the \textbf{dynamics} and the \textbf{observations}, enabling forward simulation for prediction, planning, and control of the system.

Building upon insights from cognitive science and control theory, world models from MBRL formalize internal mental simulation using a learned latent state-space representation~\cite{forrester1971counterintuitive, ha2018world, ha2018recurrent, hafner2019learning, hafner2020mastering, hafner2023mastering, hafner2025training, lee2023dreamsmooth, mattes2023hieros, liu2025continual, prasanna2024dreaming, hao2025neural}.
In this framework, the operation of a world model is also distilled into \textbf{observation}, \textbf{state}, and \textbf{dynamics}.
Specifically, the \textbf{observation} at time $t$, denoted as $O_t$, captures external sensory information and can be interpreted as the system output in a control-theoretic sense. 
In our context, $O_t$ typically takes the form of high-dimensional pixel-level visual inputs, such as video frames, providing only a partial and indirect view of the underlying environment.
To bridge the gap between raw perception and internal reasoning, the \textbf{state} $S_t$ is introduced as the agent’s latent representation of the environment.
This state serves as a learned, high-dimensional generalization of the classical system state in control theory, summarizing all task-relevant information needed for prediction and decision-making.
Although its specific parameterization varies across architectures, $S_t$ consistently serves as a compact, abstract encoding of the world's state.
Finally, the \textbf{dynamics} govern the causal evolution of this state under specific interventions or actions $A_t$, which correspond to control inputs.
This evolution is modeled through a learned state transition function
\begin{equation}
    S_{t+1} = f(S_t, A_t),
\end{equation}
mirroring the state equation in classical dynamical systems.
This predictive mechanism enables the model to simulate future states by internalizing the environment's objective physical laws, thereby allowing the agent to ``foresee'' the consequences of its actions before they are executed.

Taken together, the discussions above suggest that, despite differences in domain and formalism, the operation of a world model consistently revolves around three tightly coupled components:
\begin{itemize}
\item \textbf{Observation:} This refers to the raw, perceptible data from the environment. 
In the context of video models, observations are typically high-dimensional, pixel-level inputs (video frames) that provide a partial, indirect view of the world.
\item \textbf{State:} The state is the internal representation that aims to provide a comprehensive explanation of the current environment. To establish an accurate state, a model must distill and aggregate information from a vast history of observations. Its primary goal is to retain all task-relevant variables while filtering out irrelevant noise and redundancy.
\item \textbf{Dynamics:} The dynamics govern how the state evolves over time, often under the influence of specific actions or interventions. 
By modeling the transition $S_{t+1} = f(S_t, A_t)$, the world model internalizes the underlying physical and causal regularities of the environment.
\end{itemize}

\subsection{The Operations of World Models: Estimation and Prediction}\label{sec:pre_operations}
\begin{figure}[t]
\centering
\includegraphics[width=1.0\linewidth]{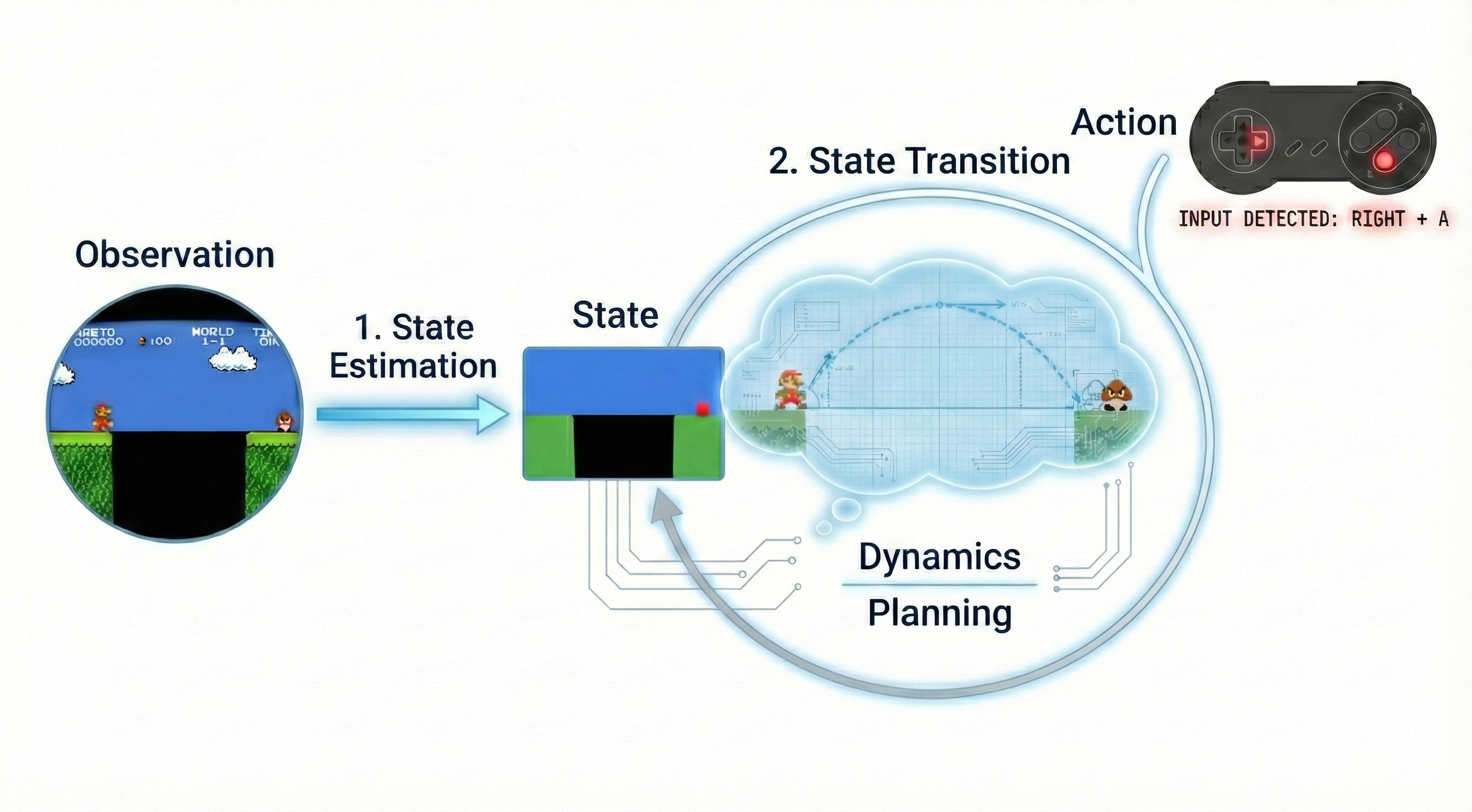}
\caption{\textbf{World Model Inference Cycle.}
\textbf{(1) Estimation:} Maps historical observations $O_{1:t}$ to a latent representation $S_t$ (Eq.~\ref{eq:state_recurrent}).
\textbf{(2) Transition:} Predicts the future state $S_{t+1}$ given current state and action $A_t$, facilitating mental rollouts and future prediction (Eq.~\ref{eq:implicit_dynamics}).}
\label{fig-method}
\end{figure}

Building upon the fundamental components introduced above, a world model operates through two core processes: state estimation and state transition.
Although their concrete implementations differ substantially across modeling paradigms, these processes share a common conceptual foundation.
In what follows, we abstract away architectural details and focus on the essential logic underlying each operation.

\vspace{0.1cm}
\noindent\textbf{1. State Estimation (Understanding the World).}
State estimation functions as the model’s abstraction mechanism. Its goal is to compress high-dimensional, sequential observations into a compact representation that captures the current status of the environment.

\begin{table*}[t]
\centering
\caption{\textbf{Comparison of World Model Paradigms.} We compare representations across Cognitive Science, Control Theory, RL, and Video Generation. The Trade-off row synthesizes strengths (\textcolor{green}{+}) and limitations (\textcolor{red}{-}).}
\small
\renewcommand{\arraystretch}{1.5} 
\setlength{\tabcolsep}{4pt}

\definecolor{goodgreen}{RGB}{0, 150, 80}
\definecolor{badred}{RGB}{200, 0, 0}

\newcolumntype{Y}{>{\centering\arraybackslash}X}

\begin{tabularx}{\textwidth}{@{} l Y Y Y Y @{}}
\toprule
\textbf{Dimension} & \textbf{Mental Model}~\cite{johnson1983mental} & \textbf{Control Theory}~\cite{kalman1960new} & \textbf{MBRL World Model}~\cite{hafner2020mastering} & \textbf{Video World Model}~\cite{openai2024sora} \\
& \textit{(Cognitive Science)} & \textit{(State-Space)} & \textit{(Latent Dynamics)} & \textit{(Generative Video)} \\
\midrule

\textbf{State}
& Abstract Concepts
& Physical Variables
& Latent Vectors
& Visual Tokens \\
\addlinespace

\textbf{Dynamics}
& Heuristic inference
& Deterministic Equations
& RNN Transition
& Global Attention \\
\addlinespace

\textbf{Observation}
& Sensory Perception
& Sensor Readings
& Compressed Visual Features
& Raw frames \\
\addlinespace

\textbf{Prediction}
& Mental imagination
& Numerical simulation
& Latent rollout
& Video generation based on history \\
\addlinespace

\textbf{Trade-off}
& \textcolor{goodgreen}{\textbf{+}} Generalization \newline \textcolor{badred}{\textbf{--}} Unverifiable
& \textcolor{goodgreen}{\textbf{+}} Precise, Interpretable \newline \textcolor{badred}{\textbf{--}} Poor Scalability
& \textcolor{goodgreen}{\textbf{+}} Planning Efficiency \newline \textcolor{badred}{\textbf{--}} Weak Interpretability
& \textcolor{goodgreen}{\textbf{+}} Real-world Alignment \newline \textcolor{badred}{\textbf{--}} Computational Cost \\

\bottomrule
\end{tabularx}
\end{table*}

Formally, this process approximates the posterior distribution of the latent state conditioned on the observation history:
\begin{equation}
S_{t} \sim P_{\phi}(S_t \mid O_{1:t})
\label{eq:state_estimation_dist}
\end{equation}

In \emph{stateful} world models, which are commonly adopted in MBRL methods~\cite{ha2018world,ha2018recurrent, hafner2019learning, hafner2020mastering, hafner2023mastering, hafner2025training}, state estimation is typically implemented through a recurrent aggregation mechanism.
The latent state is updated incrementally by integrating new observations and actions over time:
\begin{equation}
S_t = \text{Enc}_{\phi}(S_{t-1}, O_t, A_{t-1})
\label{eq:state_recurrent}
\end{equation}

This recurrent formulation enables the model to encode non-instantaneous physical properties—such as velocity, acceleration, and object permanence—which cannot be inferred from a single observation alone.

\medskip
In contrast, many recent transformer-based architectures~\cite{runway2024gen3, Bruce2024GenieGI, veo2025, openai2024sora, kuaishou2024klingai, pikalabs2024pika, decart2024oasis, valevski2024diffusion, che2024gamegen, yu2025gamefactory, he2025matrix, gao2025adaworld, huang2025vid2world, alonso2024diffusion} adopt a \emph{stateless} paradigm.
These models do not maintain an explicit recurrent hidden state. Instead, the latent representation is derived directly from a fixed temporal window of recent observations:
\begin{equation}
S_t \approx O_{t-k:t}
\label{eq:state_vlm}
\end{equation}

Here, the burden of temporal reasoning is shifted from an explicit state update to the attention mechanism operating over the observation sequence.

\medskip
At the extreme end of this spectrum lie large-scale generative video models.
Because these models operate directly on video frames—often via a Variational Autoencoder (VAE) whose latent space preserves fine-grained visual details—the notion of “state” undergoes a conceptual simplification.
In such settings, the latent state is frequently identified with the entire observation history:
\begin{equation}
S_t \equiv O_{1:t}
\label{eq:state_identity}
\end{equation}

Under this identity, state estimation reduces to maintaining the observation stream itself.
The model’s “understanding” of the world is thus fully encoded in the sequence of past perceptions, which is treated as a sufficient statistic for the current environment.
Subsequent reasoning and simulation are performed by conditioning directly on this raw or semi-raw history.

\vspace{0.1cm}
\noindent\textbf{2. State Transition (Predicting the World).}
While state estimation captures the current configuration of the environment, state transition models its causal evolution over time.
This process acts as an internal simulation engine, allowing the world model to predict future states or observations.

In stateful architectures with explicit latent representations~\cite{ha2018world,ha2018recurrent, hafner2019learning, hafner2020mastering, hafner2023mastering, hafner2025training}, transitions are governed by a parameterized latent dynamics model:
\begin{equation}
S_{t+1} \sim P_{\theta}(S_{t+1} \mid S_{t}, A_t)
\label{eq:explicit_dynamics}
\end{equation}

This formulation enforces a clear separation between representation learning and dynamics modeling, enabling efficient multi-step rollout in the latent space.

\medskip
By contrast, in stateless models without an explicit state variable~\cite{runway2024gen3, Bruce2024GenieGI, veo2025, openai2024sora, kuaishou2024klingai, pikalabs2024pika, decart2024oasis, valevski2024diffusion, che2024gamegen, yu2025gamefactory, he2025matrix, gao2025adaworld, huang2025vid2world, alonso2024diffusion}, state transition is handled implicitly through the progressive expansion of the observation sequence.

Under the identity $S_t \equiv O_{1:t}$, the dynamics reduce to an action-conditioned next-observation prediction problem:
\begin{equation}
O_{t+1} \sim P_{\theta}(O_{t+1} \mid O_{1:t}, A_t)
\label{eq:implicit_dynamics}
\end{equation}

In this formulation, “transition” corresponds to an autoregressive update of the history buffer.
By predicting the next high-fidelity observation $O_{t+1}$ and appending it to the existing sequence, the model advances the world state directly in observation space.
Through repeated application, this process effectively turns the model into a data-driven visual simulator.

\subsection{The Acquisition of World Models: Coupled vs. Decoupled Learning}\label{sec:pre_acquisition}
\begin{figure*}[t] 
\centering 
\includegraphics[width=1.0\linewidth]{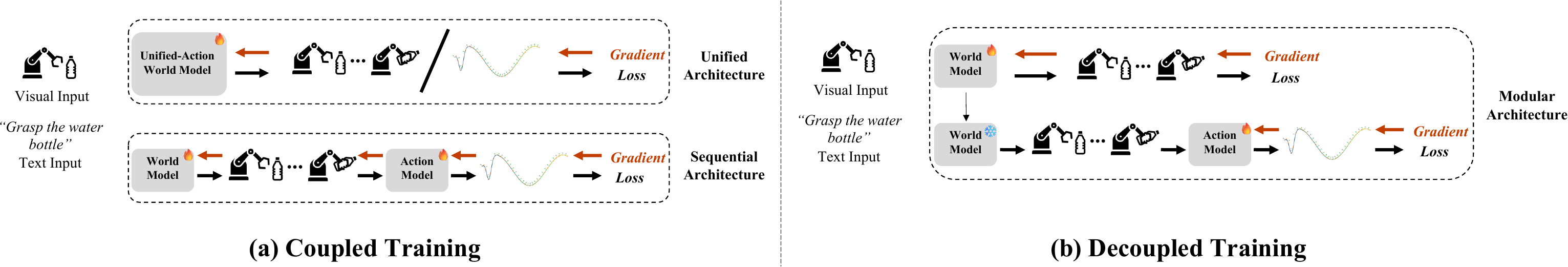} \caption{\textbf{Learning Paradigms: Coupled vs. Decoupled Training.}
\textbf{(a) Coupled Training (Closed-loop):} The world and policy models are optimized jointly via shared gradients. This often takes place in Unified Architectures or Sequential Architectures (distinct modules with continuous gradient flow).
\textbf{(b) Decoupled Training (Open-loop):} The world model serves as a pre-trained, frozen simulator. It enables policy planning but remains fixed, receiving no gradient updates from the action model.}
\label{fig-acquisition}
\end{figure*}

Since world models primarily serve to facilitate downstream decision-making, their acquisition can be categorized based on the degree of coupling with the policy model during the learning process.

\vspace{0.1cm}
\noindent\textbf{1. Closed-loop Learning (Coupled Training).} The world model and the policy model are trained jointly, such that the world model exhibits a direct gradient dependence on the policy's objectives. 
This paradigm can be further distinguished into two structural forms. 
In a \textit{sequential composition}, the models remain distinct modules but are optimized via a shared gradient flow. For instance, in ``video-as-world-model'' research such as UniPi~\cite{du2023learning} or GAIA-1~\cite{hu2023gaia}, an inverse action model is employed to extract actions from generated video sequences. 
If these extracted actions deviate from the original instructions, gradients are backpropagated into the world model to enforce ``action-executable'' and physically consistent generations. 
Alternatively, a \textit{unified architecture} integrates the world model and the policy into a single end-to-end framework, exemplified by WorldVLA~\cite{cen2025worldvla}, where perception, prediction, and action generation are optimized within a holistic system.

\vspace{0.1cm}
\noindent\textbf{2. Open-loop Learning (Decoupled Training).} They treats the world model as an independent entity learned primarily through the passive observation of large-scale datasets. 
In this paradigm, while the policy model may utilize the world model for internal ``imagination'' during its own optimization—as seen in classic MBRL frameworks~\cite{ha2018world, hafner2023mastering}—the parameters of the world model itself do not receive gradients from the policy’s reward signals or loss functions. 
Representative examples of this decoupled approach are modern video generation models~\cite{gupta2024photorealistic,lu2023vdt,ma2024latte,fan2025vchitect,liu2025lumina,zhang2025waver,wan2025wan,chen2025skyreels,ma2025step,agarwal2025cosmos,hacohen2024ltx,kong2024hunyuanvideo,genmo2025blog,openai2024sora,lin2024open,zheng2024open,yang2024cogvideox,BarTal2024LumiereAS,Chen2023SEINESV,veo2025,kuaishou2024klingai,gao2025seedance,hailuoai2025,Bruce2024GenieGI,pikalabs2024pika}. 
These models focus on mastering the objective statistical regularities of visual dynamics from vast amounts of data, providing a robust but fixed simulation substrate that remains unaffected by the specific decision-making logic of the downstream agent.

\subsection{The Evolution of Video Models: Towards Robust World Simulators}
\label{sec:pre_evolution}

Modern video generation models have emerged as a powerful candidate of the world models~\cite{openai2024sora,hu2023gaia,veo2025,du2023learning, huang2025towards,li2025worldmodelbench,kang2024far,ren2025videoworld}. 
However, despite their visual fidelity, they differ from the classical definition in several key aspects:

\subsubsection{State}
Most contemporary video generation models lack an explicit, compressed latent state. 
Instead, the observation sequence $O_{1:t}$ itself serves as an implicit state. 
However, the ever-expanding nature of these sequences imposes a significant computational burden and a growing memory footprint. 
This expansion makes long-term temporal reasoning increasingly difficult as the model must attend to a progressively larger context, often leading to a loss of persistence in long-horizon simulations. 
Consequently, recent literature focuses on enhancing \textbf{persistence} through two primary architectural strategies:

\vspace{0.1cm}
\noindent (1) \textbf{Memory Mechanisms:} These approaches maintain the implicit nature of the state within the input sequence but introduce specialized mechanisms to manage temporal data more effectively. Rather than treating all past frames equally, these methods implement operations to selectively store, retrieve, or compress elements within the sequence, thereby optimizing the model's ability to handle extended temporal contexts without linear increases in complexity~\cite{mempack2025,worldpack2025,corgi2025,xiao2025worldmem,yu2025context,song2025ditmem,an2025onestory}.

\vspace{0.1cm}
\noindent (2) \textbf{Explicit States:} In contrast, these methodologies move away from raw sequence dependence by explicitly constructing latent states that function as global, long-term memory structures. By distilling historical information into a fixed-size or hierarchical latent bottleneck, these models decouple the reasoning complexity from the sequence length, providing a stable foundation for maintaining environment consistency over distant time horizons~\cite{yu2025malt,po2025long,chen2025sana,oshima2024ssm,chen2025recurrent,yu2025videossm,wang2025lingen,gao2024matten,mo2024scaling}.

\subsubsection{Dynamics}
Standard video models~\cite{gupta2024photorealistic,lu2023vdt,ma2024latte,fan2025vchitect,liu2025lumina,zhang2025waver,wan2025wan,chen2025skyreels,ma2025step,agarwal2025cosmos,hacohen2024ltx,kong2024hunyuanvideo,genmo2025blog,openai2024sora,lin2024open,zheng2024open,yang2024cogvideox,BarTal2024LumiereAS,Chen2023SEINESV,veo2025,kuaishou2024klingai,gao2025seedance,hailuoai2025,Bruce2024GenieGI,pikalabs2024pika} often utilize bidirectional attention, generate high fidelity videos given text prompt. During the generation process, video models function primarily as ``renderers'' that produce a fixed-duration clip simultaneously, rather than exhibiting explicit temporal causality in their visual reasoning. To address the inherent lack of \textbf{causality}, current research follows two primary strategic paths to instill causality into video generation:

\vspace{0.1cm}
\noindent\textbf{(1) Causal Architecture Reformulation.} This path focuses on fundamental architectural shifts or optimization objectives to ensure the model respects temporal order. By training or distilling video models to be truly autoregressive or by employing causal masking mechanisms, researchers aim to transform the generation process from a simultaneous ``rendering'' task into a sequential ``forecasting'' task~\cite{cui2025self, liu2025rolling, sun2025ardiff, teng2025magi, deng2024nova, yu2025videomar, zhuang2025video, lin2025autoregressive, gu2025long, kondratyuk2023videopoet, deng2024autoregressive, yuan2025lumos, weng2024art, chen2024diffusion, yin2025causvid}.

\vspace{0.1cm}
\noindent\textbf{(2) Causal Knowledge Integration.} This path leverages external models that inherently possess strong causal reasoning capabilities, such as Large Multimodal Models (LMMs), to guide the generation process. This integration typically manifests in two forms: 
(1) Sequential Decoupling: A high-level LMM serves as a ``planner'' to determine the logic and dynamics of the scene, while the video model acts as a downstream ``visualizer'' to render the pixels~\cite{huang2024owl, lin2023videodirectorgpt, song2024llama, lian2023llm, yang2025vlipp, spyrou2025causally, wei2025univideo};
(2) Unified Coupling: The capabilities of understanding and generation are fused within a single framework~\cite{deng2025emerging} where the causal reasoning of the LMM and the generative power of the video model are optimized jointly to ensure that the produced dynamics are grounded in logical world knowledge.

\subsubsection{Summary}
Regarding other dimensions, such as the learning paradigm, current video models predominantly rely on Open-loop Learning, where the model is optimized independently of specific agent policies. While a significant area of active exploration focuses on whether joint training with policy-level signals is essential to achieve higher grounding and action alignment, such investigations into coupled learning remain beyond the scope of this survey. Instead, this survey focuses on investigating the state-of-the-art in video generation as a substrate for world modeling. Specifically, we review recent advancements in state representation—addressing challenges in persistence and memory—and causal dynamics—addressing the transition from rendering to physical forecasting—to chart a comprehensive technical roadmap toward more robust world simulation.

\section{Categorization - State}\label{sec:videowm_state}
To investigate the evolution of video generation models toward becoming robust world models, we first analyze their internal representations, specifically focusing on the construction of the state. As introduced in the Preliminaries, our primary objective is not necessarily to compel the model to produce an explicit state variable, but rather to incorporate the concept of a "state" as a sufficient statistic. By distilling historical context into such a representation, we ensure that the model can maintain coherent, long-term simulations. In the following sections, we will delve into two distinct paradigms—Memory Mechanisms and Explicit Latent States—to analyze how they respectively address the challenges of long-term consistency and physical persistence.

\subsection{Implicit State - Memory Mechanism}\label{sec:videowm_state_implicit}

\subsubsection{Definition}
In this paradigm, the video generation model does not explicitly construct a compact, abstract latent variable to represent the world. 
Instead, it relies on implicit state construction, where the ``state'' is functionally equivalent to a managed context of historical observations.

Here, the state $S_t$ is not a fixed-size vector but a dynamic collection of information derived from the history of observations $O_{1:t}$. 
Crucially, $S_t$ is not directly equivalent to the raw history (which would be computationally intractable) but is the result of an external \textbf{Memory Mechanism ($\mathcal{M}$)}. 

This mechanism constitutes the functional backbone of the implicit state construction. 
Drawing parallels to cognitive memory systems~\cite{sherwood2004human}which rely on encoding to compress information, consolidation to stabilize storage, and retrieval to access relevant pasts, we can formalize $\mathcal{M}$ as a composite mechanism addressing the computational constraints of video generation through three functional primitives:
\begin{itemize}
    \item \textbf{Compression:} Given the high spatio-temporal redundancy in video streams, raw history is inefficient to store. The mechanism compresses the context $O_{1:t}$ into compact representations (e.g., via token merging or summary vectors) to retain high-density information.
    \item \textbf{Retrieval:} Not all historical information is equally relevant for generating the next frame. The mechanism employs retrieval strategies (e.g., sparse attention or key-value lookup) to selectively access specific segments of the past state based on the current generation intent.
    \item \textbf{Consolidation:} Upon generating new content, the memory state must evolve. The update mechanism dictates how new observations are integrated and which obsolete information is evicted or re-weighted, enabling the model to support indefinite streaming.
\end{itemize}

Under this definition, the ``state'' represents the \textbf{Active Working Memory} of the video model, the specific subset of compressed and retrieved history required to generate the next frame.

\subsubsection{Memory Mechanism - Compression}

In the context of video generation, \textbf{Context Compression} involves condensing long historical sequences into compact representations to mitigate computational bottlenecks. While these methods often prioritize inference efficiency over explicit dynamics modeling, they serve as a critical foundation for scaling world models. By reducing the quadratic complexity of attention mechanisms, compression enables the exploration of scaling laws over extended temporal horizons. We categorize these mechanisms into \textbf{Dynamic Compression} and \textbf{Static Compression} based on the timing of the compression and whether the strategy is content-adaptive or pre-defined.

\noindent\textbf{Dynamic Compression} refers to the real-time merging or pruning of context during the model's computation, where compression criteria are typically derived from semantic similarity or attention scores. This approach has been extensively validated in LLMs through methods such as ToMe~\cite{bolya2023tome}, AdaptMerge~\cite{islam2025adaptmerge}, TokenFusion~\cite{wang2022multimodal}, MCTF~\cite{lee2024multi}, and DiffRate~\cite{chen2023diffrate}. In the video domain, VidToMe~\cite{li2024vidtome} applies these principles by merging similar tokens across frames, enhancing temporal consistency while reducing the workload to allow self-attention modules to handle longer sequences. Furthermore, works like Sparse VideoGen (SVG)~\cite{xi2025sparse} leverage an online profiling strategy to capture dynamic sparse patterns, effectively compressing the attention calculation graph by skipping non-essential spatial or temporal paths in real-time. Since the redundancy is identified based on mid-run feature activations, these methods are often training-free and highly adaptive to specific video content.

\noindent\textbf{Static Compression} involves condensing the context according to pre-defined or heuristic strategies before the model's core computation begins. While LLM-based works like Learning to Compress~\cite{mu2023learning}, RAPTOR~\cite{sarthi2024raptor}, and Long~\cite{zhang2024long} utilize specialized structures to manage long-range dependencies, video-centric methods explicitly leverage spatial-temporal redundancies. FramePack~\cite{zhang2025packing}, Pyramidal Flow Matching~\cite{jin2024pyramidal}, and LoViC~\cite{jiang2025lovic} design compression ratios based on temporal importance, where the degree of downsampling varies across the timeline, typically maintaining higher fidelity for recent frames and aggressive compression for distant history. RELIC~\cite{hong2025relic} adopts empirical spatial downsampling on the KV cache to accommodate longer horizons, whereas TempoMaster~\cite{ma2025tempomaster} implements a coarse-to-fine temporal strategy where the context is compressed via a low frame rate "blueprint" before being refined at higher rates. Because these strategies alter the input distribution or representation density, they generally require the model to be trained or fine-tuned to adapt to the compressed latent context.

\begin{figure*}[t]
    \centering
    \includegraphics[width=1.0\linewidth]{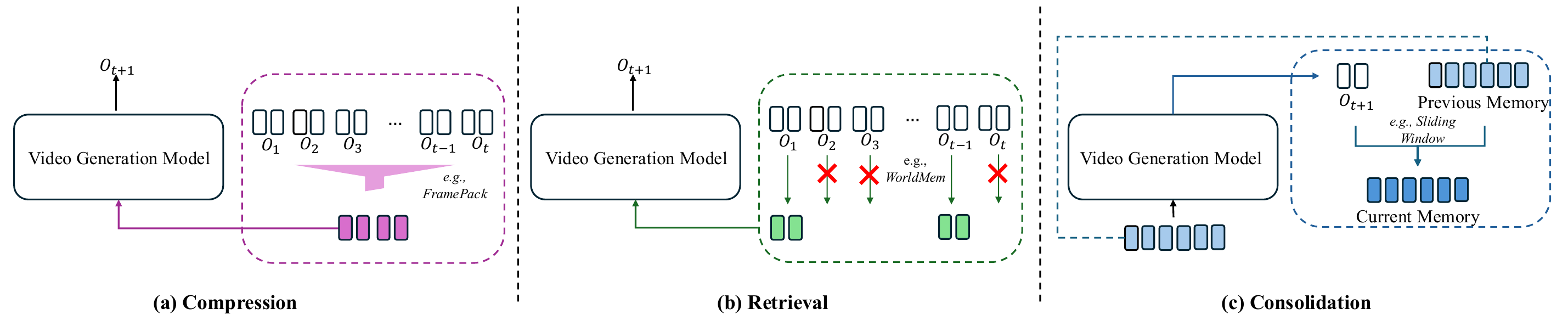}
    \caption{\textbf{Functional Primitives of the Memory Mechanism.} 
To manage the computational constraints of long-term video generation, the implicit state is governed by three operations: 
\textbf{(a) Compression} mitigates bottlenecks by condensing the raw observation history $O_{1:t}$ into compact representations.
\textbf{(b) Retrieval} prioritizes contextual relevance by selectively accessing specific historical segments via internal routing or external matching. 
\textbf{(c) Consolidation} follows a constant computational strategy by dynamically updating the memory buffer, integrating newly generated observations $O_{t+1}$, and evicting obsolete history for continuous streaming.}
    \label{fig-method}
    \vspace{-0.2in}
\end{figure*}

\subsubsection{Memory Mechanism - Retrieval}
\noindent\textbf{External Retrieval} involves the pre-selection of historical or external context before it is fed into the generative model, focusing on \emph{contextual relevance} rather than just computational throughput. This paradigm is generally divided into two categories: maintaining self-consistency through historical memory and enhancing quality via external reference-augmentation. To preserve spatial-temporal consistency, which is often referred to as an \emph{implicit geometry memory}, methods like WorldMem~\cite{xiao2025worldmem}, Context-as-Memory~\cite{yu2025context}, VRAG~\cite{chen2025learning}, and WorldPack~\cite{worldpack2025} query historical frames using low-dimensional physical states such as camera poses, 3D coordinates, and orientation. While Ctrl-World~\cite{guo2025ctrlworld} simplifies this by sampling historical frames to reduce context length, MagicWorld~\cite{li2025magicworld} utilizes the first frame of an I2V sequence as a visual key to retrieve relevant data from the history cache. 
Beyond self-history, another branch of research focuses on reference-augmented generation to inject "World Knowledge" or motion priors. Corgi~\cite{corgi2025} utilizes T2I-generated candidates to guide the model, while DiT-Mem~\cite{song2025ditmem}, RAGME~\cite{ragme2025}, and MotionRAG~\cite{motionrag2025} leverage CLIP-based embeddings to retrieve external video clips or "motion exemplars" that provide realistic physical dynamics. Finally, OneStory~\cite{an2025onestory} advances this by introducing a learnable retrieval module, demonstrating that external controllers can evolve from passive search engines into active, intent-aware components of the world simulator by optimizing the retrieval policy end-to-end.

\noindent\textbf{Internal Retrieval} implements retrieval \textbf{implicitly within the attention mechanism}. 
Instead of managing the context buffer externally, these methods integrate the ``retrieval'' process directly into the attention calculation phases. To address the computational bottleneck of long-context modeling, a prominent paradigm reframes the world state as a dynamically retrieved context, where the internal memory is managed through learnable or prior-driven visibility operators. Early approaches in this category, such as MoC~\cite{cai2025mixture} and MoBA~\cite{lu2025moba}, implement learnable routing engines that partition the history into content-aligned or contiguous chunks, utilizing mean-pooled key descriptors for top- block selection; while MoC emphasizes content boundaries like shots and frames, MoBA focuses on the mathematical consistency of merging these routed outputs via Online-Softmax. To preserve global awareness often lost in sparse selection, VSA~\cite{zhang2025faster} introduces a geometry-aligned dual-pathway architecture that fuses fine-grained local retrieval with a gated coarse global context, whereas VMOBA~\cite{wu2025vmoba} tailors this multi-scale perspective for diffusion models through a 1D-2D-3D layer-wise recurrent partition and global thresholding. Diverging from purely learned routing, AdaSpa~\cite{xia2025training} and Sparse VideoGen~\cite{xi2025sparse} leverage structural priors, specifically attention stability across denoising steps and head-specific spatiotemporal roles, to achieve training-free inference-time sparsification. The granularity of retrieval is further refined in MoGA~\cite{jia2025moga} and SVG2~\cite{yang2025sparse}, which transcend physical block boundaries by clustering tokens into semantically coherent groups or memory blocks via k-means centroids, ensuring high-fidelity identity preservation across disjoint segments. Efficiency and robustness are addressed in BSA~\cite{zhan2025bidirectional}, which proposes bidirectional pruning of both query and key-value redundancies, and ReSA~\cite{sun2025rectified}, which safeguards memory integrity by introducing periodic dense rectification to counter cumulative approximation errors and memory drift. Finally, the retrieval logic becomes increasingly specialized in Video-XL-2~\cite{qin2025video}, which employs a task-aware mixed-resolution strategy via bi-level KV representations, and Radial Attention~\cite{li2025radial}, which imposes a physics-driven distance prior to log-scale memory access based on the natural exponential decay of attention energy.










\begin{table*}[t]
\centering
\caption{Taxonomy of Memory Mechanisms in Implicit State. We categorize the management of historical context into three functional primitives: Compression, Retrieval, and Consolidation.}
\small
\renewcommand{\arraystretch}{1.6} 

\newcolumntype{Y}{>{\centering\arraybackslash}X}

\begin{tabularx}{\textwidth}{@{} c Y Y Y @{}} 
\toprule
\textbf{Feature} & \textbf{Compression} & \textbf{Retrieval} & \textbf{Consolidation} \\
\cmidrule(r){2-2} \cmidrule(lr){3-3} \cmidrule(l){4-4}

\textbf{Core Logic} 
& Condense raw history via token merging/pruning 
&  Filter context via heuristic rules
& Dynamically update buffer \& evict obsolete data \\
\addlinespace

\textbf{Cost (Time/Space)} 
& \textbf{Low / Low} 
& \textbf{Medium / Low}
& \textbf{Low / Constant} \\
\addlinespace

\textbf{Trade-off} 
& \begin{tabular}[t]{@{}c@{}}
  \textcolor{green}{+} Computational Efficiency \\ 
  \textcolor{red}{-} Lossy (Details dropped)
  \end{tabular}
& \begin{tabular}[t]{@{}c@{}}
  \textcolor{green}{+} Contextual Relevance \\ 
  \textcolor{red}{-} Rigid Rules
  \end{tabular}
& \begin{tabular}[t]{@{}c@{}}
  \textcolor{green}{+} Infinite Streaming \\ 
  \textcolor{red}{-} Semantic Drift
  \end{tabular} \\
\addlinespace

\textbf{Examples} 
& ToMe~\cite{bolya2023tome}, FramePack~\cite{zhang2025packing},  VidToMe~\cite{li2024vidtome} 
& WorldMem~\cite{xiao2025worldmem}, Context-as-Memory~\cite{yu2025context}
& StreamingT2V~\cite{henschel2025streamingt2v},  FreeLong~\cite{lu2024freelong}, EgoLCD~\cite{zhang2025egolcd} \\

\bottomrule
\end{tabularx}

\label{tab:implicit_mechanisms}
\end{table*}

\subsubsection{Memory Mechanism - Consolidation}
Memory Consolidation refers to the real-time distillation of the context buffer following the generation of new content to preserve long-range consistency while maintaining a constant computational footprint. \textbf{StreamingT2V}~\cite{henschel2025streamingt2v} implements this by extracting features from the final 8 frames of a generated chunk as a short-term window while anchoring the entire sequence to the first frame's global semantics. \textbf{FreeLong}~\cite{lu2024freelong} utilizes a spectral blending operation, using 3D FFT to isolate and merge global low-frequency structural layouts with local high-frequency details from the most recent window. For token-based modeling, \textbf{Loong}~\cite{wang2024loong} employs a pixel-to-token re-encoding of the final 5 frames to reset distribution shifts before discarding previous history. In contrast, \textbf{FAR}~\cite{gu2025long} utilizes asymmetric patchify kernels to aggregate distant historical frames into low-resolution tokens within the KV cache. \textbf{WorldWeaver}~\cite{liu2025worldweaver} expands the consolidation modality by archiving generated depth cues into a structured memory bank to constrain geometric drift. Finally, \textbf{EgoLCD}~\cite{zhang2025egolcd} introduces an active update policy through importance-driven sparse KV compression, filtering out redundant tokens based on attention scores to retain only high-value semantic features.

\subsubsection{Summary: Functional Primitives of Implicit State}

In synthesizing the landscape of implicit state construction, we categorize these approaches based on their operational logic and functional roles in managing the trade-off between computational constraints and historical fidelity (see Table~\ref{tab:implicit_mechanisms}). 
While all three primitives serve to \textbf{mitigate the bottleneck of processing raw history} ($O_{1:t}$), they fundamentally differ in how they define the relevance of information. 

\textbf{Compression} functions as a mechanism for information digestion, implementing a ``soft averaging'' strategy that systematically reduces data density regardless of specific generation intent. 
In contrast, \textbf{Retrieval} acts as an active input selection filter. Rather than uniformly condensing history, it optimizes the information flow by proactively identifying and routing only the most contextually relevant segments based on heuristic or learned priors. 
Finally, \textbf{Consolidation} operates as a \textit{post-hoc} distillation process. Distinct from the pre-selection nature of retrieval, it focuses on the dynamic maintenance of the memory buffer after generation, evicting redundant information to support infinite streaming without semantic drift. 
Collectively, these components transform a static history buffer into a dynamic working memory, ensuring that the implicit state remains a refined, high-density summary of the past.

\subsection{Explicit State}\label{sec:videowm_state_explicit}
\begin{figure}[t]
    \centering
    \includegraphics[width=1.0\linewidth]{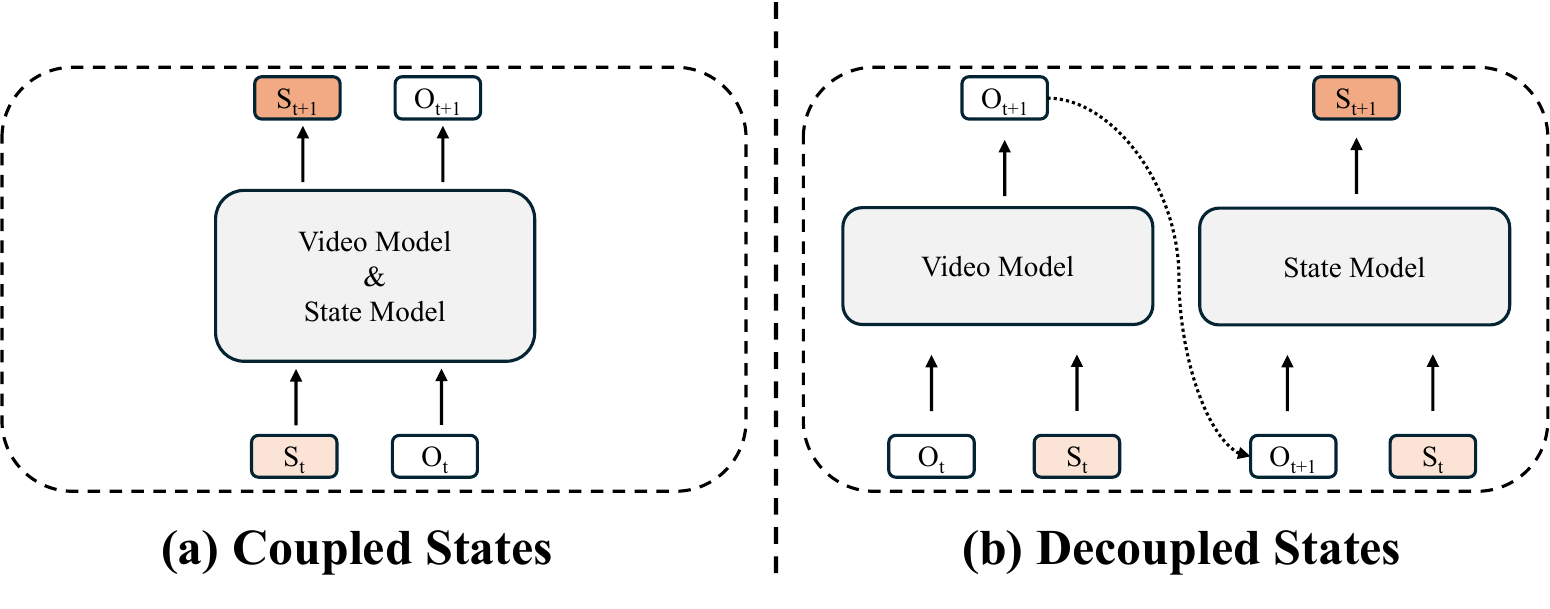}
  \caption{\textbf{Explicit State Architectures.} Instead of buffering raw history, these models maintain a compact variable $S_t$.
\textbf{(a) Coupled:} State transition is fused within the generative backbone, where $S_t$ evolves as internal hidden activations or weights.
\textbf{(b) Decoupled:} Dynamics are structurally separated; a standalone transition model updates $S_t$ before feeding it into the generator.}
    \label{fig-method}
    \vspace{-0.2in}
\end{figure}
\subsubsection{Definition}
In contrast to the ``Implicit State'' paradigm, which manages raw observations through external selection policies, this paradigm constructs the world state through internalized compression. 
Rather than maintaining a growing buffer of previous frames, these methods distill historical context into a globally updating latent variable, which serves as a comprehensive mathematical summary of the video’s evolution.

Within this framework, the ``State''  is an abstract, compact representation, ranging from transient activation vectors and parametric weights to explicit 3D geometries that evolved through a formal transition process. 
Crucially, the update mechanism shifts from heuristic rules (such as FIFO or retrieval) to a functional transition mapping, where the model learns to compress and propagate only the most salient information for future consistency. 
Depending on the architectural relationship between this transition function and the generative backbone, we categorize these works into two sub-paradigms:

\begin{itemize}
\item \textbf{Coupled States:} The transition function is intrinsically fused within the generation backbone, allowing the model to function as a synchronous state-transition system. Here, the state manifests either as \textit{hidden-variable states} (e.g., SSM hidden units, LSTM cells, or linear attention buffers) or as \textit{parametric states} where history is encoded directly into the plasticity of the model's weights through online optimization.

\item \textbf{Decoupled States:} The state is structurally decoupled from the backbone’s internal activations, existing as a standalone representation that is explicitly updated. This includes \textit{semantics-oriented} approaches that utilize separate transition models (e.g., LLMs) to evolve latent world descriptions, and \textit{geometry-oriented} approaches where the state is an explicit 3D memory (e.g., point clouds or Gaussians) updated via iterative spatial fusion and back-projection.

\end{itemize}

\subsubsection{Coupled States}
Crucially, the recursive updatability of these methods allows the observation model to function inherently as a state transition model, where the mapping and the generation of new observations are performed synchronously by a single architecture. In this paradigm, the ``memory'' of the sequence is stored within the model itself, manifesting either as hidden-variable states or parametric states.

\noindent\textbf{Hidden-Variable States.}
This class of states typically consists of hidden activation vectors maintained and updated through internal causal linear systems, such as State-Space Models (SSMs), Recurrent Neural Networks (RNNs), or linear attention mechanisms. For instance, MALT Diffusion~\cite{yu2025malt} constructs a global state by maintaining a fixed-size memory latent vector within the model layers to compress long-range context into compact tokens, using a recurrent attention mechanism to distill visual features from each generated segment. Similarly, VideoSSM~\cite{yu2025videossm} utilizes a Mamba-based SSM branch where the hidden states serve as a global compressed memory to capture dependencies beyond local sliding windows. SANA-Video~\cite{chen2025sana} leverages the cumulative properties of linear attention to maintain a constant-memory state, employing a block-wise autoregressive approach to integrate incremental features via the associative property. To resolve issues of forgetting, Recurrent Autoregressive Diffusion (RAD)~\cite{chen2025recurrent} integrates LSTM modules into the Diffusion Transformer, using hidden and cell states as memory carriers updated through an RNN forward pass. Architecture-specific scanning methods further enhance these states; LinGen~\cite{wang2025lingen} utilizes Mamba2 hidden states initialized with "review tokens" for global semantic priors and updates them via linear scanning guided by Rotary-Major Scan (RMS). Other representative works include Matten~\cite{gao2024matten}, which embeds the global state in hybrid Mamba-Attention latent variables, and DiM~\cite{mo2024scaling}, which employs bidirectional SSM modules for spatio-temporal propagation. Finally, Po et al.~\cite{po2025long} combine block-wise SSM states with local KV caches, using a spatial-major/time-minor scan order to evolve the state linearly as new frames arrive.

\noindent\textbf{Parametric States.}
This paradigm represents a fundamental shift by treating a subset of model parameters as dynamic, input-dependent variables rather than fixed constants. Instead of storing history in activation buffers, specific weights serve as state carriers that are continuously updated based on the input stream, effectively allowing the model to "learn" and solidify the dynamics of the current context on-the-fly. TTT-DiT~\cite{dalal2025one} parameterizes the global state as weight matrices within Test-Time Training (TTT) layers, where an online self-supervised gradient optimization process "writes" long-term historical information directly into the parameter space during inference. Building on this, Titans~\cite{behrouz2024titans} constructs a neural memory module that utilizes a momentum-based update rule to consider both past trends and current ``surprises'' when updating memory weights. From a theoretical perspective, Nested Learning~\cite{behrouz2025nested} defines the global state as the optimal solution to a nested optimization problem aimed at compressing the ``Context Flow'' of a sequence, distillating and solidifying context into multi-level state representations in real-time.

\noindent\textbf{Summary.} 
Hidden-variable states offer significant advantages in engineering feasibility and rapid architectural iteration, particularly through hybrid designs that combine stateless Transformer blocks with causal linear systems. However, effectively balancing state capacity with the frequency of memory updates remains a critical challenge that warrants further investigation for long-horizon modeling. In contrast, parametric states theoretically possess far greater capacity and representational power than hidden-variable states. Nevertheless, their implementation involves substantial engineering overhead and complex optimization requirements, and empirical references within the current video generation literature remain relatively nascent.

\begin{table*}[t]
\centering
\caption{Comparison of Implicit vs. Explicit State Paradigms. Implicit states prioritize visual fidelity via history buffers, while Explicit states prioritize efficiency and reasoning via compact latent variables.}
\small
\renewcommand{\arraystretch}{1.5} 
\newcolumntype{Y}{>{\centering\arraybackslash}X}
\newcommand{\multicell}[1]{\multicolumn{2}{>{\hsize=\dimexpr2\hsize+2\tabcolsep\relax}Y}{#1}}

\begin{tabularx}{\textwidth}{@{} c Y Y Y Y @{}} 
\toprule
\textbf{Feature} & \multicolumn{2}{c}{\textbf{Implicit State}} & \multicolumn{2}{c}{\textbf{Explicit State}} \\
\cmidrule(r){2-3} \cmidrule(l){4-5}
& \multicolumn{2}{c}{\textit{(Managed History / Buffer)}} & \multicolumn{2}{c}{\textit{(Condensed Variable / Latent)}} \\
\midrule

\textbf{Mechanism} 
& \multicell{External Management (e.g. Compression, Retrieval)} 
& \multicell{Internal Recurrence (e.g. $S_t \to S_{t+1}$)} \\
\addlinespace

\textbf{Context-Selection} 
& \multicell{Heuristic Learning/Rule-based selection} 
& \multicell{Learned physics/transitions} \\
\addlinespace

\textbf{Persistence} 
& \multicell{Window-Limited, Local fidelity} 
& \multicell{Global Continuity, Infinite horizon} \\
\addlinespace

\textbf{Complexity} 
& \multicell{Context-Bound: $O(N)$ or $O(N^2)$} 
& \multicell{Constant:  $O(1)$} \\
\addlinespace

\textbf{Trade-off} 
& \textcolor{green}{+}  High  Fidelity 
& \textcolor{red}{-}  High Cost 
& \textcolor{green}{+} Memory Efficient
& \textcolor{red}{-}  Information Loss \\
\addlinespace

\textbf{Examples} 
& \multicell{WorldMem~\cite{xiao2025worldmem}, Context-as-Memory~\cite{yu2025context}} 
& \multicell{VideoSSM~\cite{yu2025videossm}, RAD~\cite{li2025radial}, MALT~\cite{yu2025malt}} \\

\bottomrule
\end{tabularx}

\label{tab:state_comparison_refined}
\end{table*}

\subsubsection{Decoupled States}

Methods in this category maintain an explicit state that is structurally decoupled from the internal activations of the visual generation backbone. 
Instead of fusing dynamics into the model's weights or hidden layers, these approaches preserve a standalone state representation, ranging from abstract latent variables to explicit 3D geometry that evolves as new observations are produced. 
The mechanism of state evolution depends on the system's orientation: while semantics-oriented methods typically utilize a standalone transition model to map states across time, geometry-oriented methods treat the 3D representation itself as the state, updating it through iterative spatial fusion.

\noindent\textbf{Semantics-oriented State.}
Semantics-oriented methods focus on maintaining high-level consistency in object identity, scene logic, and narrative progression. 
By abstracting historical context into symbolic or latent semantic representations, these approaches ensure that the ``story'' of the video remains coherent over long durations, even as visual details transform. 
For instance, Owl-1~\cite{huang2024owl} implements a standalone transition model by maintaining a compact latent world state external to the video backbone; following the rendering of each clip, a multimodal LLM (MLLM) summarizes observed dynamics into a natural-language description and maps the current state to the next state to guide subsequent generation. 
Similarly, Pack and Force~\cite{wu2025pack} employs the MemoryPack mechanism, where the SemanticPack module functions as an external recurrent network. It iteratively updates a long-term memory vector  by compressing historical segments and aligning them with global text-image guidance, effectively encapsulating minute-level context within a standalone recurrent state variable.

\noindent\textbf{Geometry-oriented State.}
Geometry-oriented methods prioritize spatial consistency, view-invariance, and physical structure. A distinctive feature of this paradigm is that the 3D representation itself acts as the state, serving as a persistent, evolving memory rather than a static asset. In this framework, 3D structures, such as point clouds \cite{yu2024viewcrafter, li2025vmem, ren2025gen3c, cao2025uni3c, wu2025genfusion, huang2025voyager, wang2025evoworld}, 3D Gaussians \cite{chen2025flexworld, zhang2025scene}, meshes \cite{yang2025matrix}, or implicit fields \cite{zhai2025stargen} are progressively refined over the course of generation. Instead of relying on a learned transition model, the state is updated through direct fusion of new observations. The process typically begins with a coarse reconstruction estimated from sparse observations, followed by an iterative render–generate–update cycle: the current 3D memory is rendered from a new viewpoint to identify missing regions; a generative model synthesizes the novel view; and the synthesized image is subsequently back-projected and fused into the 3D representation to update geometry, appearance, and visibility. This iterative refinement allows the system to accumulate scene knowledge over time, ensuring rigorous structural integrity across multi-view predictions.

\noindent\textbf{Summary.} Decoupled states offer significant flexibility, allowing researchers to select specialized state spaces and transition models tailored to specific task requirements. Semantics-oriented states facilitate transitions within highly abstract LLM-driven semantic spaces, which excel at narrative logic but may lack fine-grained spatial reasoning. Conversely, geometry-oriented states leverage explicit 3D representations to ensure rigorous spatial consistency and structural integrity, though they often encounter difficulties in modeling rich and fluid temporal dynamics.

\subsection{Summary: Implicit vs. Explicit States in World Modeling}

Synthesizing the paradigms of state construction (see Table~\ref{tab:state_comparison_refined}), we analyze the divergence between implicit and explicit approaches across four critical dimensions: \textbf{Mechanism}, \textbf{Logic}, \textbf{Persistence}, and \textbf{Scalability}.

\textbf{Mechanism and Persistence.}
The fundamental distinction lies in how history is being stored. Implicit states operate as a managed History, utilizing external management mechanisms to maintain a buffer of raw observations. While this preserves high visual fidelity by keeping authentic tokens, it results in \textbf{Window-Limited} persistence, where the model is prone to forgetting once it exceeds the context window. Conversely, Explicit states employ \textbf{Internal Recurrence}, distilling history into a compact latent variable ($S_t$) through recursive updates. This formulation achieves \textbf{Global Continuity}, theoretically supporting infinite horizons, though the aggressive compression can lead to ``information decay'' and a loss of fine-grained details over time.

\textbf{Logic and Causality.}
From a causal perspective, the two paradigms define information relevance differently. Implicit states are predominantly \textbf{Heuristic-Driven}, relying on human-engineered rules (e.g., similarity matching or time proximity) to select context. In contrast, Explicit states are \textbf{Dynamics-Driven}. By requiring the model to autonomously learn the state transition function ($S_t \rightarrow S_{t+1}$), these systems move beyond pattern matching to internalize the underlying physics of the world, aligning more closely with the theoretical definition of a world simulator.

\textbf{Scalability and Trade-offs.}
Finally, computational scalability dictates the deployment feasibility. Implicit states are \textbf{Context-Bound}, with inference costs growing linearly or quadratically ($O(N)$ to $O(N^2)$) relative to history length. Explicit states, however, offer \textbf{Constant Scalability} ($O(1)$), maintaining a fixed computational footprint regardless of simulation duration. 
Ultimately, the choice represents a trade-off: Implicit states currently offer the most reliable path for high-fidelity video synthesis, while Explicit states represent the frontier for efficient, autonomous agents capable of long-term reasoning.

\section{Categorization - Dynamics}\label{sec:videowm_dynamics}
In this section, we primarily examine the sources of dynamic behavior within video generation models. To serve as effective world models for simulation, video models must possess a degree of reasoning capability. Specifically, a model must internalize underlying causal laws to ensure that its temporal rollouts remain physically plausible and logically consistent. Current research aimed at enhancing the causality of video models follows two predominant strategies.

\subsection{Causal Architecture Reformulation}\label{sec:videowm_dynamics_reformulation}
To internalize underlying causal laws and ensure physically plausible temporal rollouts, current research focuses on Causal Architecture Reformulation, primarily by re-engineering denoisers into strictly causal structures. Initial efforts in Causal-Masked Video Diffusion~\cite{sun2025ardiff, teng2025magi, deng2024nova} bridge the gap between rendering and forecasting by eliminating future-to-past information leakage. Methods like AR-Diffusion~\cite{sun2025ardiff} and MAGI-1~\cite{teng2025magi} enforce directional dependency via asynchronous or monotonic per-frame noise schedules, while Self-AR (NOVA)~\cite{deng2024nova} and VideoMAR~\cite{yu2025videomar} integrate frame-wise causal attention masks and next-frame diffusion losses to predict masked tokens based on complete preceding contexts. This autoregressive paradigm is further extended by Video-GPT~\cite{zhuang2025video}, which predicts sequential video clips through a clip-level causal mask, and FAR~\cite{gu2025long}, which uses asymmetric patchify kernels to compress distant history and maintain long-context causality. VideoPoet~\cite{kondratyuk2023videopoet} and NOVA~\cite{deng2024autoregressive} reformulate generation as next-token or next-frame prediction tasks using discrete or non-quantized tokens, while Lumos-1~\cite{yuan2025lumos} and ART-V~\cite{weng2024art} implement strict inter-frame causality through token dependency strategies and short-term context conditioning. Beyond architecture, recent techniques like Diffusion Forcing~\cite{chen2024diffusion} utilize progressive noise levels and causal masks, while CausVid~\cite{yin2025slow} employs causal distillation to improve efficiency. To mitigate exposure bias, Self-Forcing~\cite{huang2025self} and its derivatives—such as LongLive~\cite{yang2025longlive}, Self-Forcing++~\cite{cui2025self}, and Rolling Forcing~\cite{liu2025rolling}—condition the model on self-generated content or extended contexts, and Resampling Forcing~\cite{guo2025end} provides a teacher-free alternative by training on degraded context to simulate inference-time errors.

\noindent\textbf{Summary.}
The essence of these methodologies lies in integrating causal attention masks into diffusion-based frameworks and employing diversified noise-level schedules to enforce strict temporal dependency. By simulating inference-time challenges—such as error accumulation and exposure bias—through various forcing strategies, these approaches effectively mitigate the training-inference discrepancy, thereby ensuring robust physical consistency and logical plausibility during long-horizon temporal rollouts.

\subsection{Causal Knowledge Integration}\label{sec:videowm_dynamics_integration}
To enhance the causal reasoning capabilities of video models, a prominent research direction focuses on Causal Knowledge Integration, a paradigm that delegates high-level dynamics and planning to Large Multimodal Models (LMMs) while utilizing video models primarily as high-fidelity ``renderers.'' 
In these decoupled, sequential frameworks, models such as Owl-1~\cite{huang2024owl} and VLIPP~\cite{yang2025vlipp} utilize VLMs as motion planners—often via chain-of-thought reasoning—to ensure physical plausibility before the video model generates the final pixels. 
This ``director-led'' approach is further exemplified by VIDEODIRECTORGPT~\cite{lin2023videodirectorgpt}, LVD~\cite{lian2023llm}, and DirectorLLM~\cite{song2024llama}, which employ LLMs like GPT-4 or fine-tuned Llama 3 to perform complex spatiotemporal planning, structured layout generation, or human pose simulation, thereby offloading the burden of causal reasoning from the visual generator.

Moving beyond simple sequential execution, more advanced frameworks introduce tighter integration through gradient-based optimization or unified system guidance. 
For instance, CSVC~\cite{spyrou2025causally} iteratively optimizes text prompts using VLM-based textual gradients to produce causally faithful counterfactuals, while UniVideo~\cite{wei2025univideo} leverages an MLLM to guide a Multimodal DiT (MMDiT) in a unified stream for understanding and generation. Similarly, SemanticGen~\cite{bai2025semanticgen} maintains long-term consistency by decoupling the process into high-level semantic planning and low-level detail refinement stages. Ultimately, the most sophisticated implementations achieve deep architectural coupling, as seen in BAGEL~\cite{deng2025emerging}. By employing a bottleneck-free Mixture-of-Transformer-Experts architecture with shared self-attention, BAGEL unifies multimodal understanding and generation within a single system, allowing emergent world-modeling capabilities and complex reasoning to flourish through a holistic fusion of planning and simulation.


\section{Evaluation}\label{sec:eval}
Evaluating the leap from video generation to \emph{world simulation} calls for different benchmarks. Traditional metrics (e.g., IS, FVD) mainly quantify perceptual realism over short clips under passive viewing, but a usable world model must (i) produce high-quality observations, (ii) preserve a persistent state over long horizons, and (iii) obey causal/physical structure—especially under interventions. Accordingly, we organize evaluation along three core axes: \textbf{Quality} (frame-level fidelity, short-range temporal coherence, and conditioning alignment), \textbf{Persistence} (long-horizon consistency, revisitation, and memory-dependent continuity), and \textbf{Causality} (temporal reasoning, counterfactual response, and action-consistent dynamics).

\subsection{Quality}\label{sec:eval_quality}

Evaluating the generative \textit{quality} of a video model requires a multi-faceted approach that moves beyond simple per-frame realism to encompass visual fidelity, temporal dynamics, and semantic alignment. Traditional metrics like Inception Score (IS) and Fréchet Inception Distance (FID) gauge isolated frame realism, while Fréchet Video Distance (FVD)~\cite{unterthiner2018fvd} extends this to short clips to capture basic spatiotemporal consistency. However, because these metrics often miss fine-grained temporal errors or drift in longer sequences, researchers utilize specialized measures like Fréchet Video Motion Distance (FVMD)~\cite{liu2024fvmd} to penalize unnatural motion and optical flow-based metrics to ensure frame-to-frame smoothness. To unify these dimensions, comprehensive suites like \textbf{VBench}~\cite{huang2023vbench} hierarchically decompose quality into specific aspects such as motion smoothness, subject identity, and spatial relations, using pretrained models to score each automatically. This framework is further expanded in \textbf{VBench++}~\cite{huang2024vbench++}, which covers a wider range of generative tasks and introduces a \textbf{trustworthiness} dimension alongside rigorous text alignment evaluations. Text alignment is particularly critical; modern benchmarks like WorldModelBench~\cite{li2025worldmodelbench} go beyond simple CLIP similarity checks by incorporating instruction-following tests, ensuring the model not only produces realistic footage but also strictly adheres to the semantic constraints and narrative details defined in the input prompt.

\subsection{Persistence}\label{sec:eval_persistence}

A defining feature of a world simulator is \textbf{persistence}: the ability to maintain a coherent internal state over long horizons. We evaluate this capability through two primary lenses: long-horizon coherence and specific memory capacity tasks.

\subsubsection{Long-horizon Coherence}
This aspect evaluates the model's stability as the generation length increases, ensuring that the simulation does not collapse, diverge, or drift over time. Since standard metrics like FVD are typically computed on short windows and fail to capture long-term degradation, researchers employ protocols like \textbf{VBench-long}~\cite{huang2024vbench++}—an extension of VBench designed for longer videos—to monitor trends in "subject appearance change" and "background continuity" over hundreds of frames. Studies on architectures like StreamingT2V~\cite{henschel2025streamingt2v} and SANA-Video~\cite{chen2025sana} highlight the necessity of this evaluation, demonstrating that while naive models often suffer from quality collapse or identity switches after roughly 600 frames, purpose-built persistent architectures can maintain competitive FVD and consistent scene elements across minute-long sequences (around 1800+ frames).

\subsubsection{Memory Capacity Tasks}
Beyond maintaining visual consistency, persistence requires the model to "remember" specific states, causal logic, and spatial layouts, which is evaluated through distinct memory tasks. The \textbf{World Consistency Score (WCS)}~\cite{rakheja2025worldconsistencyscore} provides a holistic, no-reference metric that checks the internal logical integrity of the generated "world," specifically monitoring object permanence, relation stability, and causal compliance to ensure entities do not vanish or act erratically. To test specific environmental recall, the \textit{scene re-visitation test}~\cite{xiao2025worldmem} requires the camera to return to a previous location; performance is quantified using \textbf{reconstruction FID (rFID)}~\cite{heusel2018rFID}, where a low score indicates the model effectively retrieved the correct latent state rather than hallucinating new details. Finally, persistence is stress-tested via functional navigation tasks like the \textbf{Memory Maze}~\cite{pasukonis2022maze} or benchmarks within VR-Bench~\cite{yang2025vrbench}, where the model must generate a traversal through a complex environment without redundant loops, demonstrating an implicit spatial memory that informs future frame generation.

\subsection{Causality}\label{sec:eval_causality}

The third critical axis of evaluation is \textbf{causality}: the extent to which a generative model internalizes and adheres to the physical laws and logical progressions of the simulated environment. A robust world model must transcend mere visual plausibility to satisfy the causal constraints governing real-world events.  We categorize these evaluations into three progressive levels: \textit{reasoning}, \textit{intervention}, and \textit{planning}.

\subsubsection{Temporal Reasoning and Physical Validity}
Fundamental to causal understanding is the correct handling of event ordering and physical interactions. The \textbf{ChronoMagic-Bench}~\cite{yuan2024chronomagic} assesses long-range temporal reasoning by focusing on transformations that require strict monotonic ordering, such as biological aging or object fabrication.  Models are penalized for semantic scrambling—for instance, reverting a fully grown tree to a seedling. Metrics such as the Metamorphic Progression Score and Temporal Coherence Score quantify the realism of these temporal trajectories. High performance here indicates an implicit representation of the "arrow of time," distinguishing causal simulation from temporally agnostic video generation.

In parallel, \textbf{physical consistency} is evaluated through benchmarks like the Physics-IQ test suite~\cite{motamed2025physiq}.  This protocol conditions models on short video prompts involving deterministic physical events (e.g., collisions, fluid dynamics, gravity) and compares the generated rollout against ground truth. Accuracy is measured via four complementary metrics: Spatial IoU (localization of action), Spatio-temporal IoU (timing and location accuracy), Weighted Spatial IoU (magnitude of movement), and pixel-wise Mean Squared Error (MSE). These are aggregated into a normalized Physics-IQ score, where 100\% represents perfect physical variance replication. Current state-of-the-art models achieve scores as low as 24\%, highlighting a significant gap between surface-level visual realism and the underlying modeling of physical dynamics.

\subsubsection{Interventions and Evaluation}
A rigorous test of causal modeling is the response to \textbf{interventions}: the ability to generate coherent rollouts when initial conditions or actions are altered. Systematic tests involve generating twin sequences that diverge at a specific control point (e.g., an object is pushed vs. left stationary). A valid world model must produce distinct, logically consistent outcomes for each intervention without introducing artifacts in unrelated scene elements.

To formalize this, \textbf{World-in-World}~\cite{zhang2025world} introduces a unified platform for agent-in-the-loop evaluation.  Unlike passive open-loop benchmarks, this framework couples the video generator with an embodied agent. The world model acts as the simulator, responding to agent actions with new frames. The primary metric shifts from visual fidelity to \textbf{task success rate} across navigation and manipulation scenarios. Results indicate that visual quality does not guarantee functional utility; models with superior consistency and predictability often enable higher task success. This underscores the necessity of "intervention-centric" evaluation, where the model's value is defined by its reliability as an interactive environment.

\subsubsection{Planning and Embodied Task Performance}
The ultimate validation of a world model lies in its utility for \textbf{planning} and decision-making. Evaluations in this domain are adapted from robotics, measuring the performance of policies acting within the generated world. Key metrics include \textbf{Success Rate (SR)}. Additionally, \textbf{normalized regret} compares agent performance within the generated model against an oracle policy in a ground-truth simulator, quantifying the performance degradation attributable to model inaccuracies.

To diagnose the source of failures, researchers measure \textbf{controllability}: the alignment between the model's predicted dynamics and real-world physics under identical action sequences. Formally, for a sequence of actions $\mathbf{a}_{1:T}$, the controllability score is defined as:

\begin{equation}\label{eq:controllability}
\mathrm{Ctrl}(\mathbf{a}_{1:T}) = \exp\left( -\lambda \cdot \mathcal{D}_{\text{LPIPS}}\left( \hat{\mathbf{o}}_{1:T}(\mathbf{a}_{1:T}), \mathbf{o}^{\star}_{1:T}(\mathbf{a}_{1:T}) \right) \right),
\end{equation}

where $\hat{\mathbf{o}}_{1:T}$ represents the observation sequence generated by the world model, $\mathbf{o}^{\star}_{1:T}$ is the ground-truth sequence from a physics engine or real-world recording, and $\mathcal{D}_{\text{LPIPS}}$ is the perceptual distance metric scaled by $\lambda$. A score approaching 1 implies high fidelity to real-world dynamics. 

Finally, recent work explores emergent planning capabilities within video generation models.  For instance, VideoWorld~\cite{ren2025videoworld} demonstrates that large-scale video pretraining allows models to function as planners for complex tasks, such as the game of Go or robotic manipulation in CALVIN~\cite{mees2022calvin} and RLBench~\cite{james2019rlbench}, without explicit reinforcement learning. Success in these decision-oriented benchmarks suggests that sufficiently advanced generative models effectively internalize the causal structure of the environment, allowing them to function not just as renderers, but as actionable world models.

\section{Future work}\label{sec:future}
In this survey, we have argued that the evolution of video generation into world simulation represents a fundamental shift in the field. We identified that for current models to truly function as world simulators, they must effectively address the twin challenges of persistence and causality.

\subsection{Persistence}\label{sec:future_persistence}
Regarding persistence, future research directions diverge depending on whether the model relies on implicit or explicit state representations:

\noindent\textbf{Implicit States}. The critical imperative is to move beyond simplistic, heuristic-based memory strategies—such as fixed-length context windows—toward more advanced, data-driven memory mechanisms. These systems should leverage learned knowledge rather than hand-crafted rules, utilizing sophisticated attention-based operations to dynamically determine which information is essential for maintaining long-term consistency.

\noindent\textbf{Explicit states}. In cases such as State-Space Models, the focus should shift toward balancing computational efficiency with visual fidelity. While compact, fixed-size memory buffers offer extreme efficiency, they often act as a bottleneck for visual quality. Future work must explore hybrid strategies that preserve the efficiency of compressed states without sacrificing the fine-grained details necessary for high-fidelity simulation.

\subsection{Causality}\label{sec:future_causality}
To realize the ultimate vision of a world model, research must shift from capturing mere statistical correlations to understanding true causal mechanisms. We propose two parallel paths:

\noindent\textbf{Causal Architecture Reformulation}: This involves investigating how pre-training can bestow models with robust causal inference capabilities. This requires a fundamental exploration of model architectures and, more importantly, data preparation. Establishing sophisticated annotation methodologies and granularities to decouple latent causal factors within video data is paramount. Such decoupling is not only essential for causal reasoning but is also intrinsically linked to achieving precise control in video generation.

\noindent\textbf{Causal Knowledge Integration}: A promising short-term solution involves bridging the gap between generative and understanding models. By integrating strong reasoning priors from large-scale understanding models into the generative process, we can move toward a unified system where video creation is guided by a deep comprehension of underlying dynamics. However, effectively aligning these generative and understanding components remains a significant research challenge.

Ultimately, these advancements will enable video models to overcome the fundamental hurdles of persistence and causal consistency, bringing us closer to the realization of general-purpose world simulators.

\section{Conclusion}\label{sec:conclusion}
This survey examines the convergence of Video Generation and Model-Based Reinforcement Learning, identifying the World State ($S_t$) as the pivotal component distinguishing a simulator from a content generator.
We argue that while current architectures excel at visual synthesis, their dependence on raw observation buffers ($O_{1:t}$) imposes severe bottlenecks in reasoning and scalability.
Through our taxonomy of Implicit and Explicit paradigms, we observe a clear trajectory toward compressing historical context into compact, persistent representations.
Consequently, evaluation must shift from perceptual metrics to functional benchmarks that assess state consistency and physical dynamics.
Looking ahead, the field must transcend passive, open-loop prediction.
By enabling closed-loop interaction and causal intervention, future models will evolve from rendering pixels to simulating the governing laws of reality.

\ifCLASSOPTIONcaptionsoff
  \newpage
\fi



%

{\small
\bibliographystyle{revision_ref}
\bibliography{
ref
}
}

%

\vspace{-1cm}

\begin{IEEEbiographynophoto}{}
\end{IEEEbiographynophoto}
\vspace{-1cm}

\end{document}